\documentclass[acmlarge]{acmart}
\renewcommand\footnotetextcopyrightpermission[1]{} 
\usepackage{amsmath}
\usepackage{booktabs} 
\usepackage{algorithmic}
\usepackage{algorithm2e}
\usepackage{wrapfig}
\usepackage{graphicx}
\usepackage{multirow}
\usepackage{verbatim}
\usepackage{color}
\usepackage{subfigure}
\usepackage{array}
\usepackage{colortbl}
\usepackage{bm}
\usepackage{mathtools}
\newcolumntype{L}{>{\centering\arraybackslash}m{1.8cm}}

\acmDOI{10.475/123_4}

\acmISBN{123-4567-24-567/08/06}

\acmConference[arXiv]{}{March 2021}{Online} 
\acmYear{2021}
\copyrightyear{2021}

\acmArticle{4}
\acmPrice{15.00}

\setcopyright{none}

\begin{document}
\title{Botcha: Detecting Malicious Non-Human Traffic in the Wild}

\author{Sunny Dhamnani}
\affiliation{%
 \institution{Adobe Research}
 \country{India}
}
\email{dhamnani.sunny@gmail.com}

\author{Ritwik Sinha}
\affiliation{%
 \institution{Adobe Research}
 \country{United States}
}
\email{risinha@adobe.com}

\author{Vishwa Vinay}
\affiliation{%
 \institution{Adobe Research}
 \country{India}
}
\email{vinay@adobe.com}

\author{Lilly Kumari}
\affiliation{%
 \institution{University of Washington Seattle}
 \country{United States}
}
\email{lkumari@uw.edu}

\author{Margarita Savova}
\affiliation{%
 \institution{Adobe Systems}
 \country{United States}
}
\email{mgalabov@adobe.com}

\begin{abstract}
Malicious bots make up about a quarter of all traffic on the web, and degrade the performance of personalization and recommendation algorithms that operate on e-commerce sites. Positive-Unlabeled learning (PU learning) provides the ability to train a binary classifier using only positive (P) and unlabeled (U) instances. The unlabeled data comprises of both  positive and negative classes. It is possible to find labels for strict subsets of non-malicious actors, e.g., the assumption that only humans purchase during web sessions, or clear CAPTCHAs. However, finding signals of malicious behavior is almost impossible due to the ever-evolving and adversarial nature of bots. Such a set-up naturally lends itself to PU learning. Unfortunately, standard PU learning approaches assume that the labeled set of positives are a random sample of all positives, this is unlikely to hold in practice. In this work, we propose two modifications to PU learning that make it more robust to violations of the \textit{selected-completely-at-random} assumption, leading to a system that can filter out malicious bots. In one public and one proprietary dataset, we show that proposed approaches are better at identifying humans in web data than standard PU learning methods.
\end{abstract}

\settopmatter{printacmref=false}
\maketitle

\section{Introduction}

Non-Human Traffic or traffic generated by robots (or bots) is estimated to constitute close to half of all web traffic~\cite{badbotreport}. 
Some bots have a legitimate purpose (e.g. web crawlers) while others try to intrude the systems with malicious intent. 
It is estimated that half of all bot traffic has a malicious intent \cite{badbotreport}.
Good bots identify themselves but malicious bots have an incentive to spoof their user agents and behave like humans. 
Malicious bots may be designed to generate fake reviews, scrape price or content, crack credentials, infiltrate payment systems, defraud advertisers, or spam online forums. Recommendation and personalization systems are particularly vulnerable to bot activity \cite{zhang2006analysis}. 

The major challenge in building machine learning (ML) models to detect bad bots is getting labeled data. 
In this context, ML methods that aim to learn from positive and unlabeled data (PU learning) provide promise \cite{elkan2008learning}. 
PU learning learns from data where only a subset of one class is labeled.
We explore an application of PU learning to malicious non-human traffic detection on the web. 
Considering humans as the positive class, we can identify positive instances by assuming that only humans purchase on e-commerce web sites, clear CAPTCHAs, or visit from validated IP addresses.

\begin{figure}[t]
\centering
\includegraphics[width=0.5\linewidth]{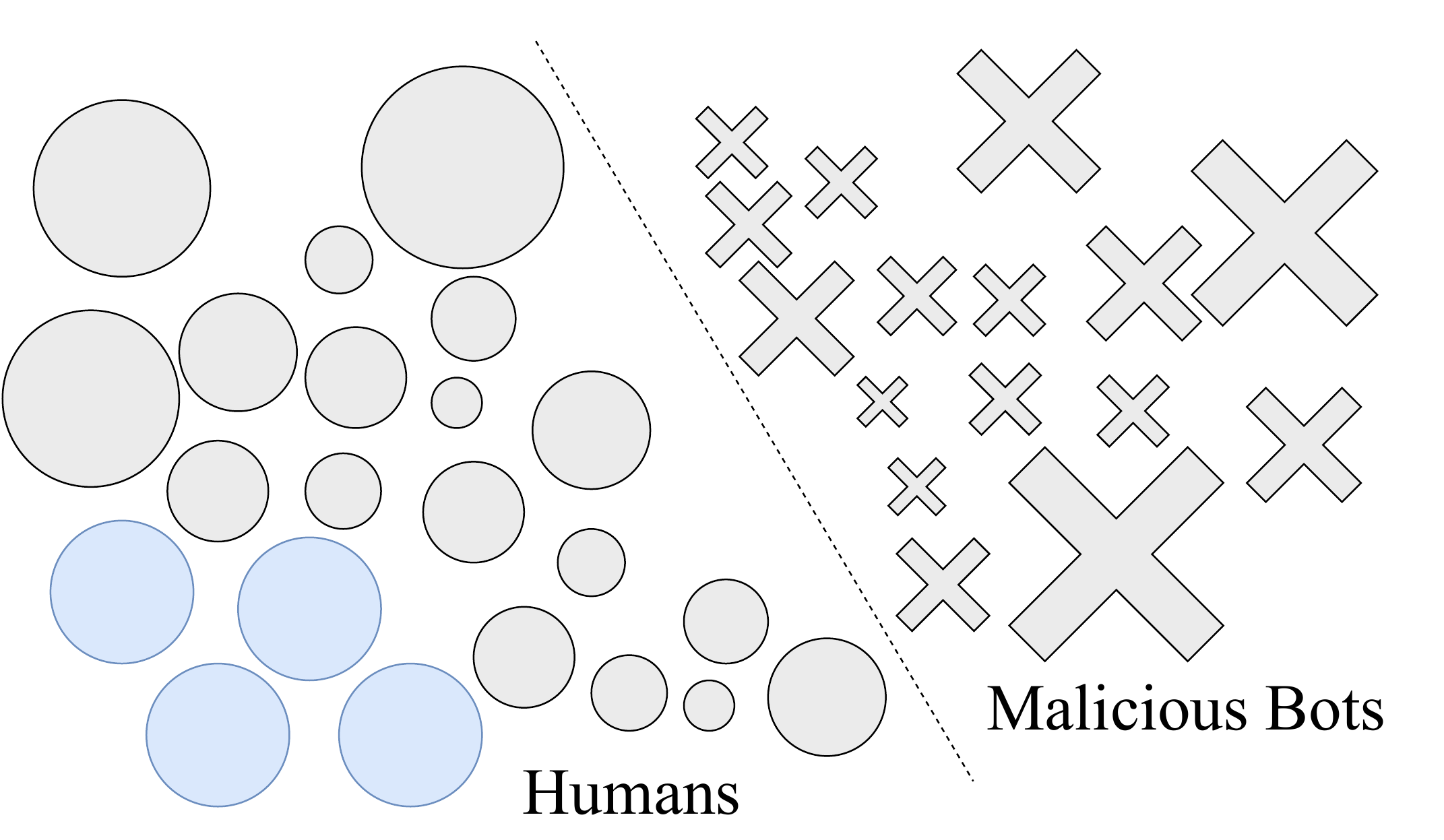}
\caption{Problem set-up: circles denote humans and crosses denote bots. Only the blue circles are known to be humans, the true label for all grey instances is unknown. The fact that larger circles are more likely to be blue denotes that the observation's attributes determine their likelihood of being selected (labeled). Under SCAR (Selected Completely at Random) violation, the classification goal is to identify the dashed dividing line between the two classes.}
\label{problem}
\end{figure}

Current PU learning frameworks assume that the labeled subset of the positive class is \textit{Selected Completely at Random (SCAR)} from the positive class, where the labeling mechanism does not depend on the attributes of the example~\cite{elkan2008learning}. That is, the labeled subset of humans is not influenced by the features of the observations. Unfortunately, such an assumption is hard to justify in practice. For example, it is reasonable to expect that not all human visitors to an e-commerce web site are equally likely to make a purchase. This requires us to revisit the PU framework to handle problems where the random sampling assumption is violated. Figure \ref{problem} describes the problem we are addressing. 

In this work, we address the question of classifying a  web session as originating from a human surfer or a robot, using PU learning. Our contribution includes two novel models to handle biased sampling within the positive class, one of which is a scalable version of the proposals in \cite{bekker2019beyond}. In our experiments, positive-unlabeled scenarios are artificially created in a publicly available intrusion detection dataset. We notice that the proposed approaches perform better than existing PU learning models~\cite{elkan2008learning, liu2003building}. In a proprietary e-commerce dataset, our methods work well in distinguishing humans from bots. We call our framework ``Botcha". Given the limited need for labeled data, it is readily applied in the wild. Filtering out all bot traffic allows recommendation and personalization systems to learn from unbiased data from real human activity.

\section{Related Work}

Malicious non-human activity on the web has been observed in the context of fake reviews, information theft, spread of misinformation, spam on social networks, and click fraud in advertising~\cite{gao2017research, displayAdFraud}. Given the diverse, dynamic and often adversarial nature of web fraud, it is imperative to find new strategies to detect bots and data driven strategies hold promise for this.

While there has been some work to build recommendation systems that are robust to adversarial attacks \cite{zhang2006analysis, he2018adversarial}, in this work we aim to filter out all bot traffic to provide unbiased data to the recommendation and personalization systems to learn from. 
To classify a visitor as a bot or human, the standard machine learning strategy requires representative examples from both classes, and building a supervised learning model that is able to differentiate between them. Due to limited labeled data for bot detection, alternative data-efficient strategies have also been investigated. Semi-supervised learning has been applied to the bot detection problem~\cite{worldThatCounts}.
Unfortunately, while it is reasonable to expect that we have a reliable subset of humans labeled, bots on the web are adversarial, ever-evolving and hard to sample from, this renders semi-supervised learning limited in scope.   

PU Learning requires only a subset of one of two classes to be labeled. Hence, PU learning is appealing in the bot detection problem, where we can assume that a subset of humans are labeled. Early work in~\cite{puWithWeightedLR} and~\cite{elkan2008learning} has shown how PU learning can achieve the effectiveness of standard supervised learning. We believe that the PU learning framework is natural for use in a variety of fraud detection applications on the web.

Empirical success in a variety of scenarios has led to  recent attention in the class of PU learning algorithms~\cite{bekker2020learning}.
Unfortunately, most prior work in this area makes the assumption that the labeled points are randomly sampled from the positive class.
This assumption is referred to as Selected Completely at Random~\cite{elkan2008learning, bekker2019beyond}. That is to say, the positively labeled examples in the dataset are a random unbiased sample of the universe of positive examples, and are not a function of the attributes of the data point. In order to allow the building of PU learning based models in scenarios where this is an unrealistic assumption, we build on prior work by \cite{bekker2019beyond}, which presents two challenges. First, the model strategy presented in~\cite{bekker2019beyond} requires the analyst to decide on a set of features to compute the propensity score. Second, the proposal requires optimisation using an Expectation-Maximization (EM) algorithm. Unfortunately, the EM Algorithm is known to be slow to converge~\cite{jollois2007speed}, given that we would like to apply this to a scenario with tens of millions of data points and hundreds of features, this presents certain challenges in the direct application of this to our work. To test our proposed algorithms, we first conduct a series of simulation experiments on standard supervised learning datasets representing different fraud-like setups - we artificially hide the true labels which we then hope to recover via the learning algorithm, thereby showing the viability of our methods.

\section{Models}
\label{sec:model}
We first describe the notations and then briefly review PU learning work in~\cite{elkan2008learning} (Section \ref{sec:vanilla}). In section \ref{sec:mam} and \ref{sec:ram}, we describe the proposed approaches, which are the main contributions of the paper.

\subsection{Notation \& Prerequisites}
\label{sec:notations}
To distinguish humans from bots we need to learn a classifier that generate the probabilities $p(y|\bm{x})$, where $y{\in}\{0, 1\}$ denotes if the observation was generated by a human ($y{=}1$) or bot and $\bm{x}$ is the feature vector.
The dataset for PU learning are examples $(\bm{x}, y, s)$ from a space $\mathcal{X} \times \mathcal{Y} \times \mathcal{S}$, where  $\mathcal{X}$ and $\mathcal{Y}$ denote the feature space and label space. The binary variable $s$ represents if the example is labeled. 
Since only positive examples (humans) are labeled, $p(y{=}1|s{=}1)=1$. Marginalizing $p(s{=}1|x)$ over $y$, we get:

\begin{equation*}
\label{marginalizeY}
p(s{=}1|\bm{x}) = p(s{=}1|y{=}1,\bm{x}) \times p(y{=}1|\bm{x}) +  p(s{=}1|y=0,\bm{x}) \times p(y{=}0|\bm{x})
\end{equation*}
Now, $p(s{=}1|y{=}0,\bm{x})=0$ since only the positive examples are labeled. This leads to
\begin{equation}
\label{eq:ratioForPU}
p(y{=}1|\bm{x}) = \frac{p(s{=}1|\bm{x})}{p(s{=}1|y{=}1,\bm{x})}
\end{equation}
Equation (\ref{eq:ratioForPU}) forms the basis of all models which we describe next.

\subsection{Vanilla Model (EAM)}
\label{sec:vanilla}
The work by Elkan and Noto is based on the SCAR assumption, which assumes that the labeled positive examples were chosen uniformly at random from the universe of positive examples \cite{elkan2008learning}.
Formally this means $p(s{=}1|y{=}1,\bm{x})$ $=p(s{=}1|y{=}1)$, i.e., the sampling process is independent of $\bm{x}$. We can rewrite equation (\ref{eq:ratioForPU}) as
\begin{equation}
\label{eq:eam}
p(y{=}1|\bm{x}) = \frac{p(s{=}1|\bm{x})}{c} \text{ , with } c=\frac{1}{n} \smashoperator{\sum_{\bm{x} : y{=}1}}P(s{=}1|\bm{x}).
\end{equation}

The constant $c$ represents the fraction of labeled positive points and $n$ is the size of the labeled set. Note that the numerator can be obtained by training a classifier that separates the labeled ($s{=}1$) points from the unlabeled ($s{=}0$). Similarly, $c$ can be estimated using this trained classifier and a validation set. Averaging predicted scores of known positives in the validation set gives estimate for $c$. 
We refer to this model as \textit{Elkan's Assumption Model} (EAM) in our experiments and it forms the baseline for our methods. For detailed derivation and discussion we refer the readers to \cite{elkan2008learning}.

\subsection{Modified Assumption Model (MAM)}
\label{sec:mam}
The SCAR assumption above enables the building PU Learning models for a range of scenarios. However, we believe that this is an unrealistic assumption and argue that explicitly accounting for selection bias for the known positives allows us to build models that are more aligned to the data.

We propose \textit{Modified Assumption Model} (MAM), geared towards practical cases where labeling is performed via a stratified procedure. Instead of using the SCAR assumption, we make a more lenient assumption that known positives come from two sub-groups, where for one the sampling depends on $\bm{x}$ and the other is independent of $\bm{x}$.

We introduce a new binary variable $b{\in}\{0, 1\}$ that indicates which of the two sub-groups a given labeled example (i.e. $s{=}1$) comes from. 
So, $b{=}0$ indicates that value of $s$ is independent of $\bm{x}$, whereas $b{=}1$ implies that value of $s$ is dependent on $\bm{x}$. Marginalizing over b, we get: 
\begin{equation}
\begin{split}
p(s{=}1|y{=}1,\bm{x}) = p(s{=}1|{y=}1,b{=}1,\bm{x}) \times p(b{=}1|y{=}1,\bm{x}) \\ +  \; p(s{=}1|y{=}1,b{=}0,\bm{x}) \times p(b{=}0|y{=}1,\bm{x}) \nonumber
\end{split}
\end{equation}

Since $s$ is independent of $\bm{x}$ when $b{=}0$, so $p(s{=}1|y{=}1,b{=}0,\bm{x}) = c$ and given that $p(b{=}0|y{=}1,\bm{x}) = 1 - p(b{=}1|y{=}1,\bm{x})$, we can re-write above equation as
\begin{equation}
\label{eq:mam_equation}
p(y{=}1|\bm{x}) = \frac{p(s{=}1|\bm{x})}{c + p(b{=}1|y{=}1,\bm{x}) \times (1{-}c)} \text{, with } c=\frac{1}{n}  \smashoperator{\sum_{\text{\textbf{\textit{x}} : y{=}1 \& b{=}0}}}P(s{=}1|\bm{x})
\end{equation}

Similar to EAM the numerator can be obtained by training a classifier that separates the labeled ($s{=}1$) points from the unlabeled ($s{=}0$). The denominator model can be trained using $b{=}1$ and $b{=}0$ sets (note that points in these sets are labeled and positive, i.e., $s{=}1$ and $y{=}1$).
The constant $c$ can be estimated by averaging the scores predicted by numerator model for examples with $b{=}0$ in the validation set. If $p(b{=}1|y{=}1,\bm{x})=0$ for all data points, i.e., sampling is independent of $\bm{x}$, we recover EAM from MAM. Our MAM proposal closely relates to the proposals made in \cite{bekker2019beyond}, however the algorithm in \cite{bekker2019beyond} does not scale to large scale bot detection datasets.

\subsection{Relaxed Assumption Model (RAM)}
\label{sec:ram}

The most general model, referred as \textit{Relaxed Assumption Model} (RAM), does not make any assumption about $s$ being independent of $\bm{x}$.
Instead, we attempt to model this process explicitly, i.e., we build a model for $p(s{=}1|y{=}1,\bm{x})$ - the denominator in equation (\ref{eq:ratioForPU}). We first acquire a set of positive unlabeled examples of $y{=}1$ with $s{=}0$ and then utilize standard binary classification methods to distinguish $s{=}0$ from $s{=}1$ amongst the positive examples.

We propose the use of a nearest neighbour based method that finds points in the dataset that are \textit{close} to the known positives but are not in the sampled set ($s{=}1$). Since any point outside the sampled set is $s{=}0$, a nearest neighbour to a $(y{=}1, s{=}1)$ point not in this set is implicitly taken to be $(y{=}1, s{=}0)$. Note that this assumption may not always be true. As in the other models, our aim is to find techniques that are robust even when the modeling assumption may be wrong. It is important to note that we do not alter the numerator in equation (\ref{eq:ratioForPU}) and hence training classifier for numerator remains identical to EAM.

\section{Experiments and Results} \label{Simulations}

\subsection{Simulated Experiments on a Public Dataset} 
\label{ssec2}
In our first set of experiments we artificially create PU learning datasets by hiding the ground truth labels of a labeled dataset during training.
We then evaluate the trained model on a labeled test set.
The simulations primarily involve controlling the subset of positive data points that are labeled for training, and all the other instances are unlabeled.
The simulated datasets having varying degree of ``randomness", one of the extremes is a completely random subset of positive samples (satisfying SCAR perfectly).
The other is extreme is a carefully crafted subset of positive samples where SCAR assumption is violated.

{
\begin{table*}[ht]
\small
\scalebox{0.85}{\begin{tabular}{|l|l|c|c|c|c|c|c|c|c|}
\multicolumn{2}{l}{} & \multicolumn{8}{c}{\includegraphics[ height=7mm]{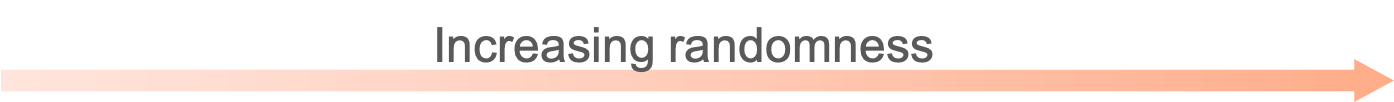}}\\
\hline
& Method & \multicolumn{2}{c|}{Mixing $m{=}0$} &  \multicolumn{2}{c|}{Mixing $m{=}30$} & \multicolumn{2}{c|}{Mixing $m{=}70$} & \multicolumn{2}{c|}{Mixing $m{=}100$}\\
& & AUC & Pr@Recall99 & AUC & Pr@Recall99 & AUC & Pr@Recall99 & AUC & Pr@Recall99\\\hline
\multirow{4}{*}[-1.5ex]{\rotatebox[origin=c]{0}{\textbf{topper $t{=}0.90$}}}
& Biased SVM \cite{liu2003building}~ & 0.705 & 0.524 & 0.705 & 0.576 & 0.688 & 0.535 & 0.689 & 0.560\\
& EAM \cite{elkan2008learning}& 0.757 & 0.719 & 0.760 & 0.751 & \cellcolor{blue!25}\textbf{0.776} & \cellcolor{blue!25}\textbf{0.751} & \cellcolor{blue!25}\textbf{0.792} & \cellcolor{blue!25}\textbf{0.697}\\
& MAM (proposed) & \cellcolor{blue!10}0.811 & \cellcolor{blue!10}0.724 & \cellcolor{blue!10}0.761 &\cellcolor{blue!10} 0.736 & 0.778 & 0.737 & 0.701 & 0.636\\
& RAM (proposed) &  \cellcolor{blue!25}\textbf{0.897} &  \cellcolor{blue!25}\textbf{0.724} &  \cellcolor{blue!25}\textbf{0.837} &  \cellcolor{blue!25}\textbf{0.756} &  \cellcolor{blue!10}0.770 &  \cellcolor{blue!10}0.743 & \cellcolor{blue!10}0.765 & \cellcolor{blue!10}0.669\\\hline

\multirow{4}{*}[-1.5ex]{\rotatebox[origin=c]{0}{\textbf{topper $t{=}0.925$}}}
& Biased SVM \cite{liu2003building}~ & 0.624 & 0.517 & 0.691 & 0.512 & 0.666 & 0.519 & 0.669 & 0.513\\
& EAM \cite{elkan2008learning}& 0.761 & 0.730 & 0.761 & 0.751 & \cellcolor{blue!25}\textbf{0.774} & \cellcolor{blue!25}\textbf{0.747} & \cellcolor{blue!25}\textbf{0.791} & \cellcolor{blue!25}\textbf{0.701}\\
& MAM (proposed) & \cellcolor{blue!10}0.831 & \cellcolor{blue!10}0.737 & \cellcolor{blue!10}0.792 &\cellcolor{blue!10} 0.752 & 0.743 & 0.717 & 0.721 & 0.682\\
& RAM (proposed) &  \cellcolor{blue!25}\textbf{0.906} &  \cellcolor{blue!25}\textbf{0.773} &  \cellcolor{blue!25}\textbf{0.812} &  \cellcolor{blue!25}\textbf{0.767} &  \cellcolor{blue!10}0.764 &  \cellcolor{blue!10}0.745 & \cellcolor{blue!10}0.748 & \cellcolor{blue!10}0.700\\\hline

\end{tabular}}
\caption {Test-set performance on public dataset. The {\color{blue!75}best algorithm} in each column is colored {\color{blue!75}blue} and {\color{blue!20}second best is light blue}. RAM and MAM perform significantly better when SCAR assumption is violated (low randomness). EAM only provides a marginal improvement over RAM when the known positives are a random subset from positive class.}
\label{tab:kdd_result}
\end{table*}

\textbf{Public Dataset:}
We use the KDDCUP'99 dataset (NSL-KDD Dataset), a widely adopted labeled dataset for network intrusion detection.
The train and test datasets have a total of $148,517$ records with $43$ features each.
To get around known problems with the dataset~\cite{kddcupCisda}, we merge the given train and test records, which we then re-split into train, validation and test sets in a 80:10:10 proportion.
Overall, the dataset contains $71,463$ intrusive sessions (all intrusions are bot-generated) while the rest are legitimate sessions.

\textbf{Data Simulations:}
The process of creating artificial datasets involves hiding the labels for all negative points and a proportion of the positive points.
We sample a labeled subset of positive data points to create the known subset of positives.
We first build a supervised classifier to score each data point.
The classification task here is to distinguish intrusive vs legitimate sessions and the score is the predicted class probability.
We use this score to introduce sampling bias in creating the known subset of positives.
Using a Random Forest classifier, we achieve an AUC (area under ROC curve) value of $0.9921$ on the training data and $0.9911$ on the test data.
We then curate different PU learning datasets by performing sampling over the scored data points by controlling two parameters as described below.

\textit{1. Topper}: This parameter is used to introduce sampling bias by selecting only those positive points whose prediction score (using supervised model) is higher than the $t^\text{th}$ quantile of all positive labeled points. 
This selection of the top fraction of positives introduces a sampling bias, since we are only selecting points with a high score.
The idea is to capture spread within the positive class, and one meaningful scale is to use the estimated probability that a point is positive, given it's features.
Note that sampling is only done for positive class, the labels for all negative points are hidden.

\textit{2. Mixing}: This parameter controls `randomness' for the known subset of positives.
After creating a sample of known positives based on topper parameter, at value $m$ we swap $m\%$ of the selected points with points from the positive set, the swapping is done with replacement.
As we move from $m{=}0$ to $m{=}100$ we decrease the sampling bias in the set and correspondingly increase the randomness. A mixing of 100\% means SCAR is completely satisfied.

The subset obtained at a particular value of $t$ and $m$ is the known labeled subset of positives, and the remaining points (all negatives and the unsampled positives) are treated as unlabeled.
With distinct values of $t$ and $m$ we obtain different simulated datasets.
At a particular value of the topper parameter ($t$), with $m{=}100$ we get a completely random sample of positive class (satisfying SCAR), on the other end with $m{=}0$ we get extremely biased sample, containing only high scoring points.
When $m{<}100$ the sampling is not completely random and depends on score of the supervised model that uses all the features $\bm{x}$.
Consequently, the sampling variable $s$ is not independent of $\bm{x}$ and the dataset does not align with the assumption of Elkan and Noto.
We show that in cases of biased sampling, the proposed methods outperform the baseline approaches that rely on SCAR assumption.

\textbf{Results on simulated datasets:}
We train MAM and RAM and compare against the baselines - EAM \cite{elkan2008learning} and biased SVM \cite{liu2003building} - on simulated datasets with varying degree of randomness.
For uniformity, we use Random Forest as the base classifier for all three methods EAM, MAM and RAM.
Biased SVM uses a SVM formulation \cite{liu2003building}.

The performance metrics are AUC (area under ROC curve) and Precision@Recall99 , precision when 99\% of known positives in the validation set are classified correctly.
Unlike the standard `0.5' threshold for classification we set classification threshold such that 99\% of the legitimate sessions (positives) are correctly classified as legitimate.
This is particularly important since in real systems we do not wish to interrupt legitimate users with any scrutiny.
And so, Precision@Recall99 is an important metric to consider.

The results for the simulated experiments are shown in Table \ref{tab:kdd_result}.
When the sampling is extreme, \textit{towards the left with smaller mixing parameter}, RAM and MAM perform significantly better than the EAM along both the evaluation metrics.
With more randomness by increasing mixing, EAM beats other methods but our proposed RAM still has a competitive performance. Biased SVM has poor performance throughout.
This shows that in extremely biased situations the proposed models MAM and RAM provide significant improvements by explicitly accounting for the sampling bias.
On the other hand, EAM provides slight improvement at high mixing (random sample) since it is tailored specifically for such scenarios, where the SCAR assumption holds true.

\subsection{Application to Real E-Commerce Data}
\label{ecomm_expt}
This section describes application of RAM to a proprietary dataset from traffic logs of an e-commerce website.

\textbf{Data Description:}
The data contains a record for every page request, here referred to as a `hit'. 
We consider a one week period, and collapse these records into `sessions' for each user.
A session combines a series of hits made by an user, the session ends with 30 minutes of inactivity.
Overall we identify $3.6$ million unique visitors from $6$ million sessions, and more than 100 million hits.
The task is to label a session as arising from a human or a bot.
The sessions from legitimate bots are filtered out using user-agent strings.
The feature representation of all sessions utilizes a standard set of technology (e.g. browser and device types), behavioral (e.g. time between hits) and session related (e.g. timezone and time-of-day).
Since this is an e-commerce website, we also have information as to whether a particular session resulted in a purchase. This information is leveraged to build our partial set of positives, the details are presented next.

\textbf{Known subset of positives:}
Out of the $6$ million sessions, $36k$ $(0.6\%)$ sessions are \textit{purchase} sessions and $360k (6\%)$ sessions belong to an identified purchaser.
We label this $6\%$ of sessions as positive (Human class).
The dataset is then split into train, test, and validation sets in a 80:10:10 ratio for modeling purposes.

\textbf{Partially labeled test dataset:}
To validate our approach, we split the test data into $3$ groups of points and observe the distribution of prediction scores across these classes. This split is based on heuristics which we describe next. \textit{Positive data points}: The subset of sessions which had a user corresponding to a purchase session in the training dataset. \textit{Negative data points}: The subset of sessions which have been originated from AWS/Azure servers are tagged as negative, the assumption being that browsing sessions originating from these cloud environments are unlikely to be initiated by humans. The set of azure/AWS IPs are publicly available \cite{18,19}. \textit{Unlabeled data points}: The set of sessions which are neither tagged as positive nor negative.

It is important to note that all points during model training had the label of `Positive' or `Unlabeled'.
The `Negative' label is only used for validation.
Note that the set of known positives is neither complete (not all Humans purchase), nor is it an unbiased sample (different users have a varying propensities for purchases). 

\begin{table}[ht]
\centering
\scalebox{0.85}{
\begin{tabular}{LLLL}
\hline
Class of sessions & No. of sessions & No. predicted human & \% predicted human\\\hline
Positive & 74k  & 73k & $\sim 99\%$\\
Negative & 24k & 608 & $\sim 2.5\%$\\
Unlabeled & 1.08M & 890k & $\sim 82\%$\\
\hline\\[-1em]
Total & 1.18M & 965k & $\sim 82\%$\\
\hline
\end{tabular}
}
\caption{Test-set observations for E-commerce dataset.}
\label{table:obs}
\end{table}

Using the validation set, we identify a threshold that captures $99\%$ of positive labels. And the output score of the RAM model is converted into a boolean \textit{is-human} label by using this threshold. Table \ref{table:obs} shows the break-up of the traffic in the dataset and how RAM classifies points from each of these classes. As seen in the table, we misclassify only a few negatively labeled sessions (<3\%) and in total, we have close to 82\% human traffic as reported by this model. We expect a high human traffic since the website has strict login requirements for accessing their content. Additionally, we observe a stark separation in the prediction scores for positive and negative class, most positive samples had a score close to 1, while negatives were scored close to 0.

\section{Conclusions}

In this paper, we have addressed the problem of detecting and filtering non-human traffic using positive and unlabeled data. Providing recommendation and personalization systems unbiased data to learn from, leads to a better experience to the end-customer. 
We specifically accounted for the \textit{selected completely at random} assumption in standard PU Learning methods and conducted simulation studies for validation.
We also evaluated our most general model, RAM, on a large real word e-commerce dataset.
We showed that the model clearly separates the known positives from negatives chosen via a heuristic.
Given the scale of fraud due to bots, such bot detection systems have a clear utility.
And the methods described in this paper show promising results in addressing the endemic bot problem.

\bibliographystyle{ACM-Reference-Format}
\bibliography{acmart} 


\begin{thebibliography}{15}


\ifx \showCODEN    \undefined \def \showCODEN     #1{\unskip}     \fi
\ifx \showDOI      \undefined \def \showDOI       #1{#1}\fi
\ifx \showISBNx    \undefined \def \showISBNx     #1{\unskip}     \fi
\ifx \showISBNxiii \undefined \def \showISBNxiii  #1{\unskip}     \fi
\ifx \showISSN     \undefined \def \showISSN      #1{\unskip}     \fi
\ifx \showLCCN     \undefined \def \showLCCN      #1{\unskip}     \fi
\ifx \shownote     \undefined \def \shownote      #1{#1}          \fi
\ifx \showarticletitle \undefined \def \showarticletitle #1{#1}   \fi
\ifx \showURL      \undefined \def \showURL       {\relax}        \fi
\providecommand\bibfield[2]{#2}
\providecommand\bibinfo[2]{#2}
\providecommand\natexlab[1]{#1}
\providecommand\showeprint[2][]{arXiv:#2}

\bibitem[\protect\citeauthoryear{AWS}{AWS}{2020}]%
        {18}
\bibfield{author}{\bibinfo{person}{AWS}.} \bibinfo{year}{2020}\natexlab{}.
\newblock \bibinfo{title}{{AWS IP} Address Ranges}.
\newblock \bibinfo{howpublished}{\url{https://amzn.to/2z2Ql7h}}.
\newblock
\newblock
\shownote{Accessed: 2020-04-27.}


\bibitem[\protect\citeauthoryear{Bekker and Davis}{Bekker and Davis}{2020}]%
        {bekker2020learning}
\bibfield{author}{\bibinfo{person}{Jessa Bekker} {and} \bibinfo{person}{Jesse
  Davis}.} \bibinfo{year}{2020}\natexlab{}.
\newblock \showarticletitle{Learning from positive and unlabeled data: A
  survey}.
\newblock \bibinfo{journal}{\emph{Machine Learning}}  \bibinfo{volume}{109}
  (\bibinfo{year}{2020}), \bibinfo{pages}{719--760}.
\newblock


\bibitem[\protect\citeauthoryear{Bekker, Robberechts, and Davis}{Bekker
  et~al\mbox{.}}{2019}]%
        {bekker2019beyond}
\bibfield{author}{\bibinfo{person}{Jessa Bekker}, \bibinfo{person}{Pieter
  Robberechts}, {and} \bibinfo{person}{Jesse Davis}.}
  \bibinfo{year}{2019}\natexlab{}.
\newblock \showarticletitle{Beyond the selected completely at random assumption
  for learning from positive and unlabeled data}. In
  \bibinfo{booktitle}{\emph{Joint European Conference on Machine Learning and
  Knowledge Discovery in Databases}}. Springer, \bibinfo{pages}{71--85}.
\newblock


\bibitem[\protect\citeauthoryear{Elkan and Noto}{Elkan and Noto}{2008}]%
        {elkan2008learning}
\bibfield{author}{\bibinfo{person}{Charles Elkan} {and} \bibinfo{person}{Keith
  Noto}.} \bibinfo{year}{2008}\natexlab{}.
\newblock \showarticletitle{Learning classifiers from only positive and
  unlabeled data}. In \bibinfo{booktitle}{\emph{Proceedings of the 14th ACM
  SIGKDD international conference on Knowledge discovery and data mining}}.
  ACM, \bibinfo{pages}{213--220}.
\newblock


\bibitem[\protect\citeauthoryear{Gao, Tang, Liu, Zhang, and Liu}{Gao
  et~al\mbox{.}}{2017}]%
        {gao2017research}
\bibfield{author}{\bibinfo{person}{Haichang Gao}, \bibinfo{person}{Mengyun
  Tang}, \bibinfo{person}{Yi Liu}, \bibinfo{person}{Ping Zhang}, {and}
  \bibinfo{person}{Xiyang Liu}.} \bibinfo{year}{2017}\natexlab{}.
\newblock \showarticletitle{Research on the security of microsoft’s two-layer
  captcha}.
\newblock \bibinfo{journal}{\emph{IEEE Transactions on Information Forensics
  and Security}} \bibinfo{volume}{12}, \bibinfo{number}{7}
  (\bibinfo{year}{2017}), \bibinfo{pages}{1671--1685}.
\newblock


\bibitem[\protect\citeauthoryear{He, He, Du, and Chua}{He
  et~al\mbox{.}}{2018}]%
        {he2018adversarial}
\bibfield{author}{\bibinfo{person}{Xiangnan He}, \bibinfo{person}{Zhankui He},
  \bibinfo{person}{Xiaoyu Du}, {and} \bibinfo{person}{Tat-Seng Chua}.}
  \bibinfo{year}{2018}\natexlab{}.
\newblock \showarticletitle{Adversarial personalized ranking for
  recommendation}. In \bibinfo{booktitle}{\emph{The 41st International ACM
  SIGIR Conference on Research \& Development in Information Retrieval}}.
  \bibinfo{pages}{355--364}.
\newblock


\bibitem[\protect\citeauthoryear{Jollois and Nadif}{Jollois and Nadif}{2007}]%
        {jollois2007speed}
\bibfield{author}{\bibinfo{person}{F-X Jollois} {and} \bibinfo{person}{Mohamed
  Nadif}.} \bibinfo{year}{2007}\natexlab{}.
\newblock \showarticletitle{Speed-up for the expectation-maximization algorithm
  for clustering categorical data}.
\newblock \bibinfo{journal}{\emph{Journal of Global Optimization}}
  \bibinfo{volume}{37}, \bibinfo{number}{4} (\bibinfo{year}{2007}),
  \bibinfo{pages}{513--525}.
\newblock


\bibitem[\protect\citeauthoryear{Lee and Liu}{Lee and Liu}{2003}]%
        {puWithWeightedLR}
\bibfield{author}{\bibinfo{person}{Wee~Sun Lee} {and} \bibinfo{person}{Bing
  Liu}.} \bibinfo{year}{2003}\natexlab{}.
\newblock \showarticletitle{Learning with positive and unlabeled examples using
  weighted logistic regression}. In \bibinfo{booktitle}{\emph{In Proceedings of
  the 20th International Conference on Machine Learning}}
  \emph{(\bibinfo{series}{ICML'03})}. \bibinfo{pages}{448--455}.
\newblock


\bibitem[\protect\citeauthoryear{Li, Martinez, Chen, Li, and Hopcroft}{Li
  et~al\mbox{.}}{2016}]%
        {worldThatCounts}
\bibfield{author}{\bibinfo{person}{Yixuan Li}, \bibinfo{person}{Oscar
  Martinez}, \bibinfo{person}{Xing Chen}, \bibinfo{person}{Yi Li}, {and}
  \bibinfo{person}{John~E. Hopcroft}.} \bibinfo{year}{2016}\natexlab{}.
\newblock \showarticletitle{In a World That Counts: Clustering and Detecting
  Fake Social Engagement at Scale}. In \bibinfo{booktitle}{\emph{Proceedings of
  the 25th International Conference on World Wide Web}} (Montr\&\#233;al,
  Qu\&\#233;bec, Canada) \emph{(\bibinfo{series}{WWW '16})}.
  \bibinfo{pages}{111--120}.
\newblock
\showISBNx{978-1-4503-4143-1}


\bibitem[\protect\citeauthoryear{Liu, Dai, Li, Lee, and Yu}{Liu
  et~al\mbox{.}}{2003}]%
        {liu2003building}
\bibfield{author}{\bibinfo{person}{Bing Liu}, \bibinfo{person}{Yang Dai},
  \bibinfo{person}{Xiaoli Li}, \bibinfo{person}{Wee~Sun Lee}, {and}
  \bibinfo{person}{Philip~S Yu}.} \bibinfo{year}{2003}\natexlab{}.
\newblock \showarticletitle{Building text classifiers using positive and
  unlabeled examples}. In \bibinfo{booktitle}{\emph{Data Mining, 2003. ICDM
  2003. Third IEEE International Conference on}}. IEEE,
  \bibinfo{pages}{179--186}.
\newblock


\bibitem[\protect\citeauthoryear{Microsoft}{Microsoft}{2017}]%
        {19}
\bibfield{author}{\bibinfo{person}{Microsoft}.}
  \bibinfo{year}{2017}\natexlab{}.
\newblock \bibinfo{title}{Microsoft Azure Datacenter {IP} Ranges}.
\newblock \bibinfo{howpublished}{\url{https://bit.ly/36aMDon}}.
\newblock
\newblock
\shownote{Accessed: 2020-04-27.}


\bibitem[\protect\citeauthoryear{Networks}{Networks}{2020}]%
        {badbotreport}
\bibfield{author}{\bibinfo{person}{Distil Networks}.}
  \bibinfo{year}{2020}\natexlab{}.
\newblock \bibinfo{title}{{2020: Bad Bot Report | IT Security's Most In-Depth
  Analysis on Bad Bots}}.
\newblock \bibinfo{howpublished}{\url{https://bit.ly/2Azqx3d}}.
\newblock
\newblock
\shownote{Accessed: 2020-05-15.}


\bibitem[\protect\citeauthoryear{Stitelman, Perlich, Dalessandro, Hook, Raeder,
  and Provost}{Stitelman et~al\mbox{.}}{2013}]%
        {displayAdFraud}
\bibfield{author}{\bibinfo{person}{Ori Stitelman}, \bibinfo{person}{Claudia
  Perlich}, \bibinfo{person}{Brian Dalessandro}, \bibinfo{person}{Rod Hook},
  \bibinfo{person}{Troy Raeder}, {and} \bibinfo{person}{Foster Provost}.}
  \bibinfo{year}{2013}\natexlab{}.
\newblock \showarticletitle{Using Co-visitation Networks for Detecting Large
  Scale Online Display Advertising Exchange Fraud}. In
  \bibinfo{booktitle}{\emph{Proceedings of the 19th ACM SIGKDD International
  Conference on Knowledge Discovery and Data Mining}}
  \emph{(\bibinfo{series}{KDD '13})}. \bibinfo{pages}{1240--1248}.
\newblock


\bibitem[\protect\citeauthoryear{Tavallaee, Bagheri, Lu, and
  Ghorbani}{Tavallaee et~al\mbox{.}}{2009}]%
        {kddcupCisda}
\bibfield{author}{\bibinfo{person}{Mahbod Tavallaee}, \bibinfo{person}{Ebrahim
  Bagheri}, \bibinfo{person}{Wei Lu}, {and} \bibinfo{person}{Ali~A. Ghorbani}.}
  \bibinfo{year}{2009}\natexlab{}.
\newblock \showarticletitle{A Detailed Analysis of the KDD CUP 99 Data Set}. In
  \bibinfo{booktitle}{\emph{Proceedings of the 2009 IEEE Symposium on
  Computational Intelligence in Security and Defense Applications (CISDA)}}.
  \bibinfo{pages}{1--6}.
\newblock


\bibitem[\protect\citeauthoryear{Zhang, Ouyang, Ford, and Makedon}{Zhang
  et~al\mbox{.}}{2006}]%
        {zhang2006analysis}
\bibfield{author}{\bibinfo{person}{Sheng Zhang}, \bibinfo{person}{Yi Ouyang},
  \bibinfo{person}{James Ford}, {and} \bibinfo{person}{Fillia Makedon}.}
  \bibinfo{year}{2006}\natexlab{}.
\newblock \showarticletitle{Analysis of a low-dimensional linear model under
  recommendation attacks}. In \bibinfo{booktitle}{\emph{Proceedings of the 29th
  annual international ACM SIGIR conference on Research and development in
  information retrieval}}. \bibinfo{pages}{517--524}.
\newblock


\end{thebibliography}

\end{document}